\documentclass[10pt,twocolumn,letterpaper]{article}

\usepackage{arxiv}
\usepackage{times}
\usepackage{epsfig}
\usepackage{graphicx}
\usepackage{amsmath}
\usepackage{amssymb}

\usepackage[breaklinks=true,bookmarks=false]{hyperref}

\arxivfinalcopy

\ifarxivfinal\fi

\usepackage{xcolor}

\begin{document}

\title{SizeGAN: Improving Size Representation in Clothing Catalogs}

\author{Kathleen M. Lewis \\
Massachusetts Institute of Technology\\
\and
John Guttag\\
Massachusetts Institute of Technology\\
}

\maketitle
\ifarxivfinal\thispagestyle{empty}\fi

\begin{abstract}Online clothing catalogs lack diversity in body shape and garment size. Brands commonly display their garments on models of one or two sizes, rarely including plus-size models. To our knowledge, our paper presents the first method for generating images of garments and models in a new target size to tackle the size under-representation problem. Our primary technical contribution is a conditional  generative adversarial network that learns deformation fields at multiple resolutions to realistically change the size of models and garments. Results from our two user studies show SizeGAN outperforms alternative methods along three dimensions -- realism, garment faithfulness, and size -- which are all important for real world use.  
\end{abstract}

\section{Introduction}
\label{sec:intro}

While brands sell multiple sizes of garments, their online stores often only show images of one or two sizes. This \textit{size under-representation problem} can be frustrating for customers who are unable to visualize what a garment might look like on someone of their size. Furthermore, studies have shown that women's health can be negatively impacted by seeing images of only thin models~\cite{body_exposure,model_body_size}. In theory, vendors could solve this size under-representation problem by hiring a more diverse set of models to represent all clothing sizes, but they rarely do so. 

\begin{figure}[ht]
\centering
\includegraphics[width=0.45\textwidth]{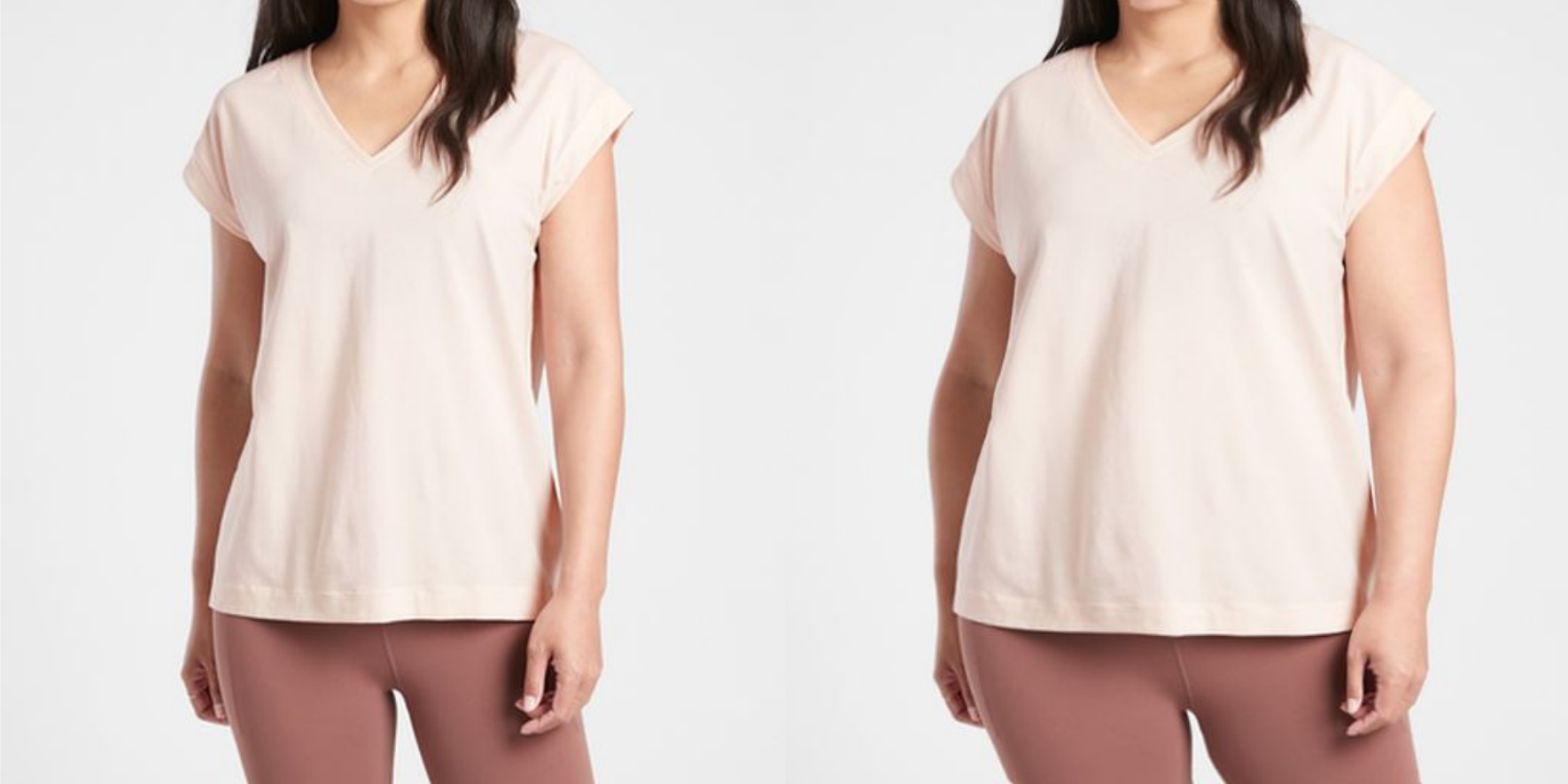}
\caption{SizeGAN takes an image of a model wearing a shirt in one size (left) as input and realistically changes the garment and model size (right).}
\label{fig:teaser}
\end{figure}

Our work aims to provide a machine learning solution to the size under-representation problem, which has been largely overlooked in the literature. We propose a generative machine learning method, SizeGAN, that, given an image of clothing worn by a model of one size, generates an image showing what that clothing would look like in a new size on a model of that size. We show that SizeGAN can be used to generate garments in an under-represented size while maintaining the details and realism of the garment. In this paper, we focus on resizing tops between sizes S/M and plus sizes 1X/2X. We show that the SizeGAN architecture is able to resize garments in both directions.

SizeGAN aims to synthesize images that are successful along three dimensions:

\begin{enumerate}
    \item \textbf{Realism}: the generated image should look like a real photo (no artifacts, unrealistic proportions, or blurred garments). Both the model and the top should look realistic enough that it could be shown in an online catalog.
    \item \textbf{Garment faithfulness}: the new-sized top in the generated image should have the same details as the original top (patterns, text, fit, shape, colors). Although the garment will be in a new size, it should visibly be the same garment.
    \item \textbf{Sizing}: the image should represent the intended size. Both the model and the garment should represent the target size and be recognizable as that size to an online shopper.
\end{enumerate}

SizeGAN uses a conditional generative adversarial network~\cite{stylegan2ada} to generate deformation fields at multiple resolutions.
Given an image of a model wearing one size of clothing, our network uses the deformation fields to change the size of the model and garment (Figure~\ref{fig:teaser}). 

To our knowledge, we are the first paper to focus on using technology to solve the size under-representation problem. Since there is no prior work suitable for direct comparison, we explore and evaluate alternative approaches to solving this problem. We evaluate our method against virtual try-on methods, which can be used to display garments on a model of a new size. We additionally compare our method to a single-axis deformation baseline, which is an alternative approach to resizing garments and models.

We evaluate our method and the try-on methods, FlowStyleVTON~\cite{flowtryon} and VITON-HD~\cite{vitonhd}, through a user study. We ask participants to rank the results along the three dimensions, realism, garment faithfulness, and sizing. We show our method outperforms existing methods on all three dimensions, with an average rank (lower is better) of 1.14 for realism, 1.17 for garment faithfulness, and 1.46 for sizing. 

We also present user study results comparing our method to the single-axis deformation baseline. We ask users to evaluate which results look most similar to the target size. When directly compared, SizeGAN was preferred 79\% of the time.

Our contributions in this paper are:
\begin{enumerate}
    \item We present a novel method, SizeGAN, that builds on contributions from StyleGAN2~\cite{stylegan2ada} and image alignment methods~\cite{Voxelmorph,stn} to address a {\it size under-representation} problem that has been overlooked by prior work.
    \item We run two user studies to compare our method with the most promising approach, using current try-on methods, and an alternative resizing approach, the single-axis deformation baseline. In both studies, our method was consistently preferred. 
    \item We make publicly available a new fashion dataset that contains images of models wearing tops with size annotations. Each top appears in two different sizes. The dataset reflects diversity in top styles, model identity, and model poses. 

\end{enumerate}

\section{Related Work}
Our work draws from two distinct fields: the use of generative adversarial networks (GANs) to produce fashion-related images and the use of deformable transforms in medical imaging. 

\subsection{Generative Adversarial Networks}

Recent works have shown that GANs are capable of producing images indistinguishable from real images in a variety of applications. StyleGAN2~\cite{stylegan2} has been shown to work well in the fashion and virtual try-on domains~\cite{PosewithStyle,TryOnGAN}. Our work builds on the StyleGAN2 architecture and uses augmentations from StyleGAN2-ADA~\cite{stylegan2ada}. Importantly, we modify the architecture to output deformation fields, instead of images, at each resolution. This change helps to preserve garment details while changing the garment shape to fit a different size person.

Virtual try-on methods aim to synthesize a person wearing a new garment. The results from these methods have improved significantly in recent years~\cite{dressinorder,CVTON,highfid,clothformer,TryOnGAN,fitgan,Trilevel}.  BodyGAN~\cite{bodyGAN} is a recent generative adversarial network method that aims to disentangle body shape, pose, and appearance. The images are compelling, however the paper does not analyze or show results for multiple sizes, and there is no public code available.  

TryOnGAN~\cite{TryOnGAN} builds an optimization method on top of StyleGAN2 to achieve photorealistic try-on. SizeGAN builds on these ideas, but differs from  TryOnGAN in task, loss function, and outputs. SizeGAN aims to change the size of a person and garment without losing details. Because TryOnGAN has to project images into the GAN latent space to work on real images, it often loses important garment details (e.g. texture, patterns, and sometimes even color). SizeGAN avoids these issues by instead learning and applying deformation fields to change the person and garment size without losing these details.

Similarly to our work, FlowStyleVTON~\cite{flowtryon} uses the StyleGAN2 architecture to learn dense flow fields. Rather than the sizing problem, their method learns coarse and refined flow fields to better handle large misalignments between garments and people. As we show, when applied to the sizing problem, the trained augmentation network does not generalize to new sizes and does not retain details from the original garment.

VITON-HD~\cite{vitonhd} is another virtual try-on methods that demonstrates high quality results for a diverse set of garments and models. This method accepts as input higher resolution images than other virtual try-on methods, however, it still does not produce catalogue quality images of clothing in diverse sizes (e.g., clothing details are not preserved).

Synthesizing diverse clothing sizes is not a direct motivation for try-on methods. Still, their ability to synthesize models wearing garments suggests that they might be applied to the resizing problem. However, our experiments indicate that current methods do not generalize well to generating new sizes. 

\begin{figure*}[t]
\centering
\includegraphics[width=0.9\textwidth]{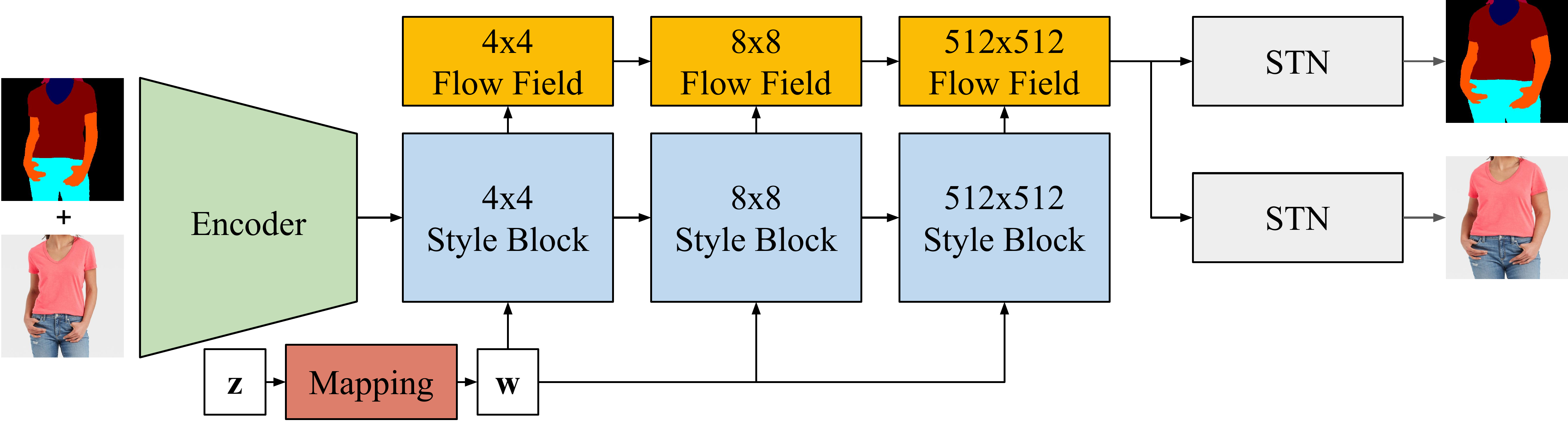}
\caption{The SizeGAN generator is conditioned on an input image, $I^A$, and segmentation, $S^A$, of a person wearing garment $x$ in size $A$. The generator also takes in a latent vector, $z$, which is mapped to an intermediate latent vector, $w$. The generator outputs a deformation field (also known as a flow field) at every resolution. These deformation fields are upsampled and combined. A spatial transformer network (STN) warps the conditional image and segmentation using the final deformation field. This changes the size of the garment and the model to size $B$ in the output image, $\tilde{I}^{B}$, and output segmentation, $\tilde{S}^{B}$.}
\label{fig:method}
\end{figure*}

\subsection{Deformable Transforms}
Deformable transforms, optical flow, and affine alignment are all methods commonly used in medical imaging and in other fields to change the geometry of an image, often for template matching. We gain insight from these methods as a way to change the size of a conditional garment while preserving the details of the garment. 

Optical flow estimation outputs a dense displacement vector for a pair of images and is often used for video frame alignment. These methods frequently require ground truth displacement vectors~\cite{depth,raft}, which can be difficult to acquire. We therefore focus on affine and deformable transforms that have been show to work in an unsupervised setting~\cite{Voxelmorph}. 

Affine alignment and deformable transforms are commonly used in the medical imaging literature to align, or register, a pair of medical images. Often affine alignment is used first as a pre-processing step and then a deformable transform, which has more degrees of freedom, is applied. The deformation fields (also referred to as flow fields) give a displacement vector for each pixel (or voxel) in an image. VoxelMorph~\cite{Voxelmorph} provides a fast, unsupervised deep learning method for finding deformation fields that align pairs of medical images. SizeGAN builds directly on ideas from VoxelMorph and uses deformation fields to change the size of a garment and model.  

\subsection{Human Body Reshaping}
Similar to our work, \cite{humanBody} uses deformation fields to change a person's appearance. Their motivation is for photo retouching, whereas our method focuses on changing garment sizing. They do not provide training code, and furthermore, their method requires ground truth retouched images for training and ground truth deformation fields derived from the ground truth images. In the size under-representation problem, we do not have ground truth images of models in the same pose in different sizes, and therefore cannot use \cite{humanBody} as a baseline. 

The parametric model approach~\cite{parametric2010} requires fitting a 3D model to each image and additionally requires manual user adjustments per image. This is not practical at scale. Our work provides a solution that is unsupervised (no paired ground truth images or ground truth deformation fields) and is practical at scale. 

\section{Method}
Given an image, $I^A$, of a person wearing garment $x$ in size $A$, we aim to generate an image, $\tilde{I}^{B}$, of the person and the garment in size $B$. We develop a conditional generative method, SizeGAN, that learns to warp the conditional image, $I^A$, such that the person and clothing matches the size $B$ distribution, while remaining realistic. Our method learns the size $B$ distribution from data through our loss functions. We train the network using our new public dataset\footnote{Our dataset will be released once this paper is published.} containing pairs of ground truth images, $I^A$ and $I^B$, that show each garment, $x$, in sizes $A$ and $B$ on different people.  

\subsection{SizeGAN Architecture}
Our architecture (Figure~\ref{fig:method}) builds on StyleGAN2~\cite{stylegan2} and StyleGAN2-ADA~\cite{Karras2020ada}, which has been shown to generate high quality images, even when the training set is small. We make several key modifications as described in the following two subsections.  

\subsubsection{Conditional GAN}
To make the generator conditional, we replace the constant $4\times4$ input to the StyleGAN generator with a $4\times4$ encoding of our conditional input, $I^{A}$, and its segmentation, $S^{A}$. We compute $S^{A}$ from $I^{A}$ using an existing method~\cite{Graphonomy}. Our encoder architecture is based on TryOnGAN~\cite{TryOnGAN} and consists of downsampling convolution layers.

\begin{figure}[t]
  \centering
 \includegraphics[width=0.7\linewidth]{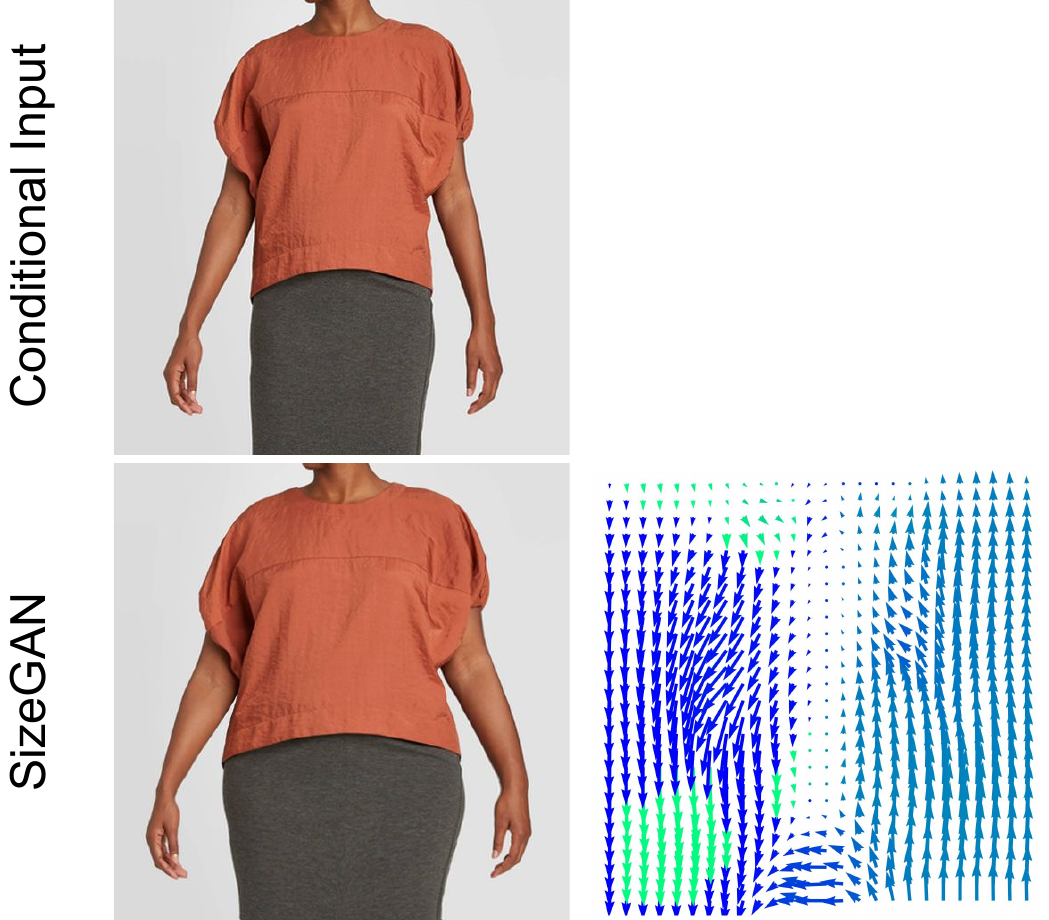}
  \caption{The left column shows the conditional image, $I^A$ and the resized garment image, $\tilde{I}^{B}$, from SizeGAN. The right column visualizes the final deformation field from our method. For better visualization, we use a stride of 20 to reduce the number of plotted points in each deformation field.}
  \label{fig:flowfields}
\end{figure}

\subsubsection{The Deformation Field} Deformation fields have been used in many applications to change the geometry of an image while preserving important features. Deformation fields (also known as flow fields) give a displacement vector for each pixel in an image. In our method, we use this displacement vector to change the size of the model and garment without losing any of the garment details (patterns, stripes, text, etc.). Instead of learning to synthesize an image directly, we learn a deformation field that is used to warp the conditional image and segmentation into the final image and segmentation. The generator outputs a deformation field at each resolution (from $4\times4$ up to the final resolution $512\times512$). At each resolution, the deformation field is upsampled and added to the deformation field generated at the next resolution. The final deformation field is used to warp $I^{A}$ into $\tilde{I}^{B}$ as well as warp the segmentation $S^{A}$ into its corresponding $\tilde{S}^{B}$. The deformation field is applied to the image and segmentation using a spatial transformer network~\cite{Voxelmorph,stn}. An example of a learned deformation field is shown in Figure~\ref{fig:flowfields}.

\begin{figure}[b]
  \begin{center}  \includegraphics[width=\linewidth]{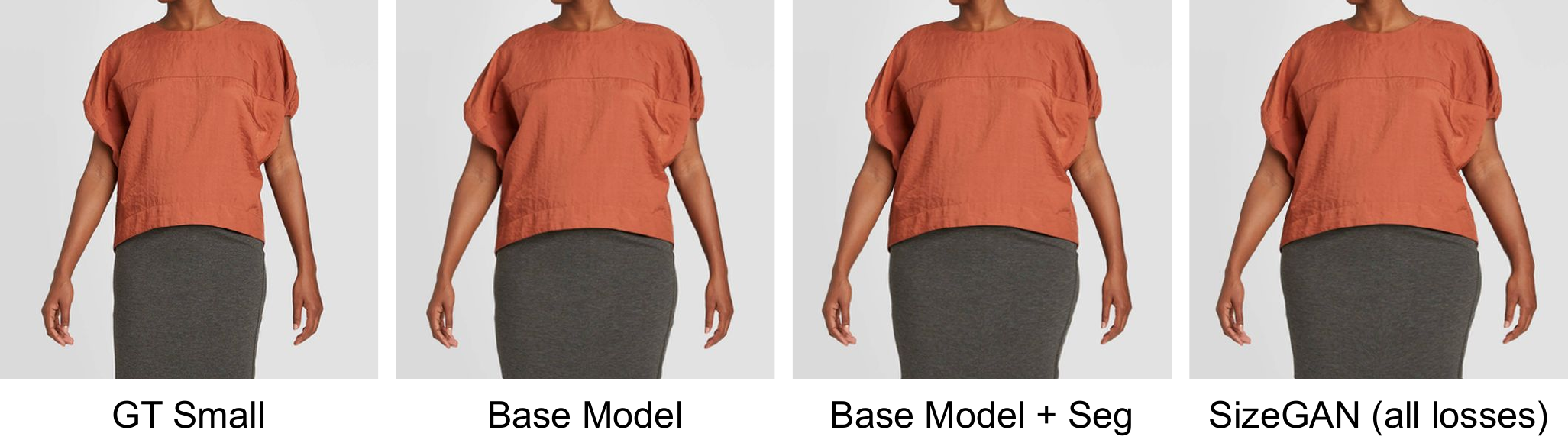}
  \end{center}
   \caption{Ablation study showing the importance of each image matching loss term. The base model includes an image discriminator. Adding the segmentation discriminator and adding the classifier further improves the results. }
   \label{fig:ablation}
\end{figure}

\subsection{Learning Objectives}
Building on the VoxelMorph method~\cite{Voxelmorph}, we attempt to learn a deformation field during training that aligns two images, $m$ and $f$, by solving an optimization of the form:

\begin{equation}\label{VMlossRelatedWork}
    \mathit{L(f,m,\phi)} = \mathit{L_{sim}(f, m\circ\phi)} + \lambda \mathit{L_{smooth}(\phi)},
\end{equation}
where $\phi$ is the deformation field, $m$ $\circ$ $\phi$ is $m$ warped by the deformation field, $\lambda$ is a regularization hyperparameter, $L_{sim}$ is an image matching term, and $L_{smooth}$ is a regularization or spatial smoothness term. The smoothness term in SizeGAN penalizes the local spatial gradients of the displacement. This constrains the deformation field to be spatially smooth such that neighboring pixels move together. This improves 1) semantic consistency between the original image and the output image, and 2) physical realism. 

The SizeGAN image matching term differs significantly from previous work. In comparison to VoxelMorph, which requires a \textbf{target image} to guide deformations, SizeGAN learns deformations to match a \textbf{target distribution}. Because many different body types can wear the same size, we do not constrain our deformations to change the body shape to match a particular template. Our method does not limit the body transformations to one type, but rather is guided by matching the target garment size distribution. Instead of using an image similarity loss such as L2 or cross correlation, SizeGAN introduces three distribution matching loss terms: 1) an image discriminator, 2) a segmentation discriminator, and 3) a binary cross entropy loss. We show in Figure \ref{fig:ablation} that each image matching loss term improves the resulting image.

The image discriminator takes either $I^B$ or $\tilde{I}^{B}$ as input and encourages the generator to produce deformation fields that produce realistic images. The segmentation discriminator takes pairs, $(S^A,S^B)$ or $(S^A,\tilde{S}^B)$, of size $A$ and size $B$ segmentations as input and guides the network to learn deformation fields that change each segment in a realistic way. The binary cross entropy loss pushes the network to produce final images that are classified as the target garment size. We pre-train a ResNet50 classifier~\cite{resnet} to distinguish between sizes and incorporate the classifier into our network loss. This encourages the network to learn deformation fields that push the synthetic image, $\tilde{I}^{B}$, closer to the correct size distribution. When training the generator, we freeze the weights of the classifier and use the predictions to calculate a binary cross entropy loss. 

\subsection{Dataset}\label{sec:ourDataset} To train our network to map between two size distributions, we collect a new dataset, which we will make publicly available. We collect women's clothing images from two brands, Target and Athleta, that have plus sized model images in addition to images of smaller clothing sizes. The dataset focuses on women's tops and includes a range of shirt styles from blouses to athletic clothing. The dataset contains pairs of $512\times512$ images of tops in different sizes. Each pair consists of a top in either a small or medium size and the same top in a 1X or 2X plus size. Although the paired images contain the same top, the paired images differ in many aspects such as pose, styling of top (tucked in or paired with accessories), background differences, and lighting differences. We collect 2,053 pairs and randomly split these image pairs into 1870 training pairs, 116 validation pairs, and 67 test pairs. We used the validation images to choose hyperparameters, and show results on the test set in the experiments.

\section{Experiments}\label{sec:experiments}

In this section, we provide implementation details. We then present results from two user studies comparing SizeGAN to existing related work. These experiments demonstrate that SizeGAN outperforms other methods on both in-distribution data from our held-out test set and on out-of-distribution data from another dataset. We provide additional results in the supplement. 

\begin{figure}[t]
  \centering
  \includegraphics[width=0.9\linewidth]{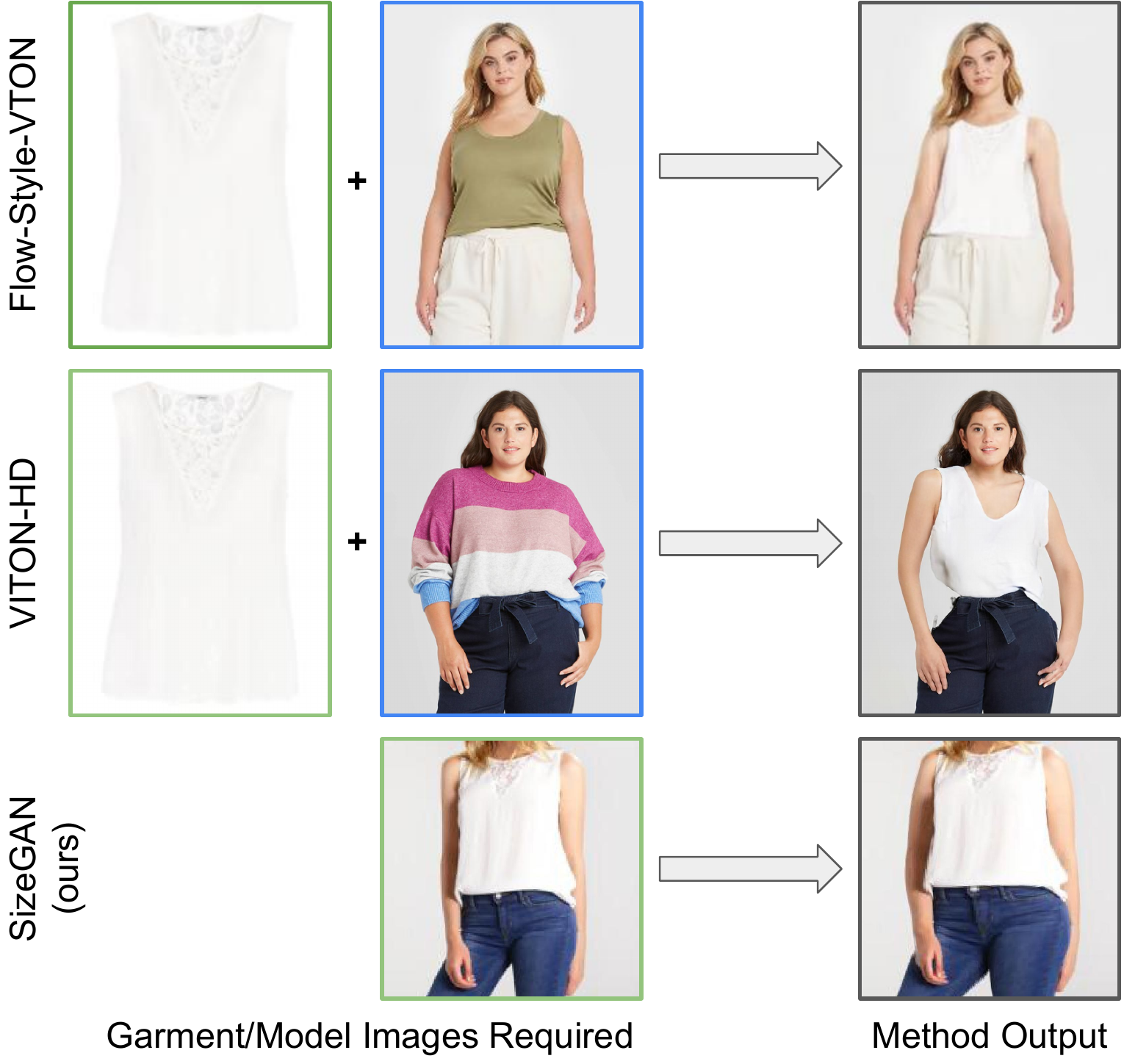}
  \caption{The rows show the garment/model image inputs required for SizeGAN and two virtual try-on methods. The VITON dataset (green bordered images) provides images of only tops and images of the tops on models (usually small-sized models). Our dataset (blue bordered images) provides model images for models of different sizes. We can apply virtual try-on methods to the task of resizing garments by combining the garment images from VITON and plus sized model images from our dataset.}
  \label{fig:inputs}
\end{figure}

\subsection{Implementation}
Our GAN model is derived from the Pytorch~\cite{Pytorch} version of StyleGAN2-ADA~\cite{Karras2020ada}. We use the default augmentations for StyleGAN2-ADA. For SizeGAN, we derive the spatial transformer network starting from the implementation of VoxelMorph~\cite{Voxelmorph}. We compute segmentations~\cite{Graphonomy} and pose keypoints~\cite{8765346,cao2017realtime,simon2017hand,wei2016cpm} for each training image. We group the computed segmentation labels into 9 coarser segments (background, upper garments, lower garments, accessories, face, hair, arms, legs, torso skin) as done in TryOnGAN~\cite{TryOnGAN}. We use the pose keypoints to automatically crop the images and then resize to $512\times512$. Since the focus of this work is resizing tops, we crop each image to show the model from chin to above the knee.

For our binary cross entropy loss term, we pre-train a size classifier (which classifies the size of the top) on our training set with size labels. When training the classifier, we use horizontal flips and color jitter as data augmentations. Our classifier is trained from scratch on $224\times224$ images with a batch size of 64. The size classifier has an accuracy of 93.5\% on our validation set. The misclassified examples either had difficult poses or low contrast between the garment and the background (e.g. white shirt against white background). 

We used the validation set of our dataset to find the best weights to balance the loss terms. We tried a large range of values for each loss weight, including zero, and found the best weights to be $30$ for the deformation field regularization loss and $1,000$ for the binary cross entropy loss. The two discriminators are weighted equally with a loss weight of one. We used the random seed 117 for sampling latent vectors to produce our results. We trained our model for $15.5$ hours using four Titan Xp GPUs.

\subsection{Baselines}
Since we are tackling a new problem, there are no direct baselines to compare against. Instead, we provide analysis of other plausible approaches to this problem. 

\textbf{Try-On Baselines} While the resizing task is not directly in-scope for try-on methods, it is a plausible extension and is the best existing approach we found to compare against our method. Virtual try-on methods can be applied to the garment resizing task by transferring the garment to a person of the desired size. We compare our method to two recent virtual try-on methods~\cite{vitonhd,flowtryon}. For both methods, we use the pre-trained models. VITON-HD~\cite{vitonhd} does not provide training code for re-training. Their model was trained on their own collected dataset\footnote{This dataset recently became public, but was not public when we ran our experiments.}. Flow-Style-VTON~\cite{flowtryon} was trained on the VITON dataset and we use their provided test split of this dataset for inference. Flow-Style-VTON has two available sets of pre-trained weights, one for in-distribution images and one for out-of-distribution images. We use their out-of-distribution pre-trained model because the input model image is from our dataset (Figure~\ref{fig:inputs}). 

\textbf{Single-axis Deformation Baseline} The single-axis deformation baseline is an alternative approach to resizing. For the single-axis deformation approach, we take the average distance between hip pose keypoints in each size distribution in the training set and scale $I^A$ by this ratio (1.36) in the x direction to get $\tilde{I}^{B}$. We show results for additional ratios in the supplement. 

\subsection{Comparison with Existing Methods}\label{sec:dataset}
Our goal is to make progress towards increasing size representation in online clothing stores. Consequently, we designed a user study to shed light on how online shoppers, who scroll through and observe images of models wearing garments, would perceive the generated images. 
\subsubsection{Dataset} We use the VITON dataset~\cite{VITON} because it contains garment images required for try-on methods. The VITON dataset has image pairs of the garment and of the garment worn on a model (usually a small-size model). The images are $256\times192$ resolution. The virtual try-on methods require garment images (obtained from VITON dataset) in size $A$ and images of size $B$ models (obtained from our dataset). SizeGAN requires only an image of a size $A$ model wearing the target garment (Figure~\ref{fig:inputs}). To compare the same garment, $x$, across all three methods, we use the model images from VITON as input to our method. We were unable to re-train the baselines because of a lack of training code or a lack of paired garment-only/model data, which we could not find for plus size clothing. However, we make this comparison more fair by also evaluating our method on out-of-distribution data since we do not train our method on the VITON dataset. 

\begin{table*}[tb]
  \renewcommand{\arraystretch}{1.5}
  \begin{center}
  \begin{tabular}{|c|c|c|c|}
    \hline 
    Methods & Realism & Garment Faithfulness & Sizing \\
    \hline\hline
    VITON-HD~\cite{vitonhd} & 2.35 (0.583) & 2.62 (0.548) & \textbf{1.92 (0.643)} \\
    Flow-Style-VTON~\cite{flowtryon} & 2.51 (0.592) & 2.21 (0.589) & 2.62 (0.623)\\
    SizeGAN (ours) & \textbf{1.14 (0.395)} & \textbf{1.17 (0.433)} & \textbf{1.46 (0.699)} \\
    \hline
  \end{tabular}
  \end{center}
  \caption{Average rank results from our user study comparing SizeGAN to VITON-HD~\cite{vitonhd} and Flow-Style-VTON~\cite{flowtryon} on the garment resizing task. We evaluate the results along three dimensions: photorealism, faithfulness to the original garment, and sizing. For each of these dimensions, a ranking of 1 indicates the best method and a ranking of 3 indicates the worst method. For each column, we state the average rank with the standard deviation in parentheses. }
  \label{tab:1}
\end{table*}

\begin{figure}[tb]
  \centering \includegraphics[width=0.9\linewidth]{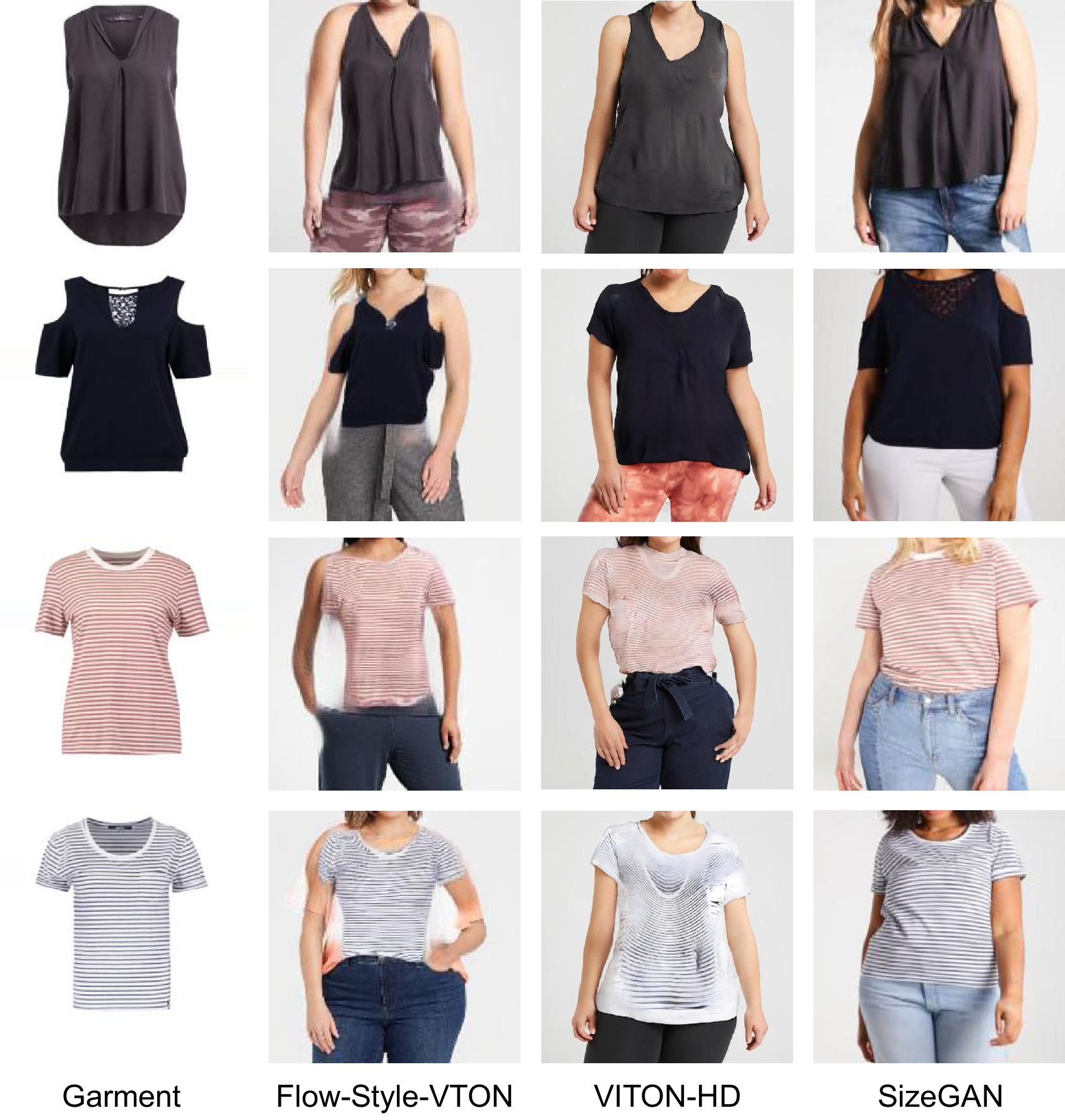}
  \caption{In comparison to state-of-the-art virtual try-on methods, SizeGAN preserves the garment details better and generates a more realistic image when resizing the garment to a larger size. For example, in the third and fourth row, the original garment has stripes that are not preserved in the two virtual try-on results. }
  \label{fig:tryon}
\end{figure}

\subsubsection{User Study Setup} Our user study focuses on synthesizing larger sizes from smaller sizes. We had $43$ participants from Amazon Mechanical Turk.  Each were asked to evaluate $15$ examples from each method. We provide the user study link\footnote{User study link: https://forms.gle/3qNYysQxv9PEWphv7} for readers to view the survey given to the participants. Each question showed the target garment on the left and the results from all three methods on the right, in a random order. Participants were asked to rank each example from: 
\begin{enumerate}
    \item most realistic to least realistic
    \item most faithful garment representation (patterns, text, shape, colors) to least faithful garment representation 
    \item largest size to smallest size model
\end{enumerate}
The three synthetic results were labeled as ``A", ``B", and ``C". Participants were allowed to say any or all of the images are equivalent. We included three questions asking users to explain their answer as quality control. These short answer questions were asked for the first example, last example, and an example in the middle of the survey. If the participant's free-form explanation was not consistent with their ranking choice, we threw those responses out. We collected a total of 53 survey responses, but 10 were thrown out based on this criteria. We also examined the short answers to get insight into strengths and weaknesses of the methods.

The images shown in the survey were randomly chosen using the following process. We first randomly selected 15 garment images from the VITON test set (more details in the supplement). For the two try-on methods, we randomly select 15 plus-sized models for each method from our dataset. We ensured all of the chosen models had in-distribution poses following the guidelines mentioned in the Dataset section. We show example inputs for each method in Figure \ref{fig:inputs}.

We resized the model images and VITON garment images to the correct image resolution for each method. After we had all the results, we cropped each image to the chin and mid-thigh so that all method images look similar. We resized all images to $512 \times 512$.

\subsubsection{User Study Results}
Table~\ref{tab:1} shows the average rank results calculated from the 43 participants in our user study. Our method clearly outperforms the baselines on realism and garment faithfulness. Our method and VITON-HD both perform well on size, whereas Flow-Style-VTON seemed to not alter the garment size. VITON-HD, while able to resize the garment, is not able to maintain garment details and realism (Figure~\ref{fig:tryon}). Similar to the online shopping experience, the users saw multiple images of the same garment and had to make quick visual-based decisions about the garment appearance and fit. 

In addition to rankings, the user study participants provided free-form comments for three of the user study examples. The actual free-form comments referred to the images as A, B, or C. When showing comments here, however, we have replaced the letter by the name of the method that generated the image. We provide all survey responses in the supplement.

The first survey question asked users to explain their rankings for realism. The set of images shown for this question appears in the bottom row of Figure~\ref{fig:tryon}. The responses gave us insight into why participants found each image as a whole believable or unbelievable. For example, one participant said ``The pattern and contours of the shirt are virtually identical in image [SizeGAN]. [Flow-Style-VTON] has the pattern of the original sized shirt but is lost on the sides and [VITON-HD] barely looks like the original shirt."

When participants were asked to explain their rankings for the garment faithfulness, users commented more specifically on the resized garment. Our results were again preferred over the baselines (Table~\ref{tab:1}), however there was more variety in which details were important to people's decision making. Some users paid attention to fit as part of garment realism and others ignored the fit of the garment, focusing only on the pattern details and shape. Furthermore, the level of detail in responses differed across participants, which could indicate that our results capture both snap-decision impressions and thoroughly analyzed opinions. Figure~\ref{fig:tryon} shows visual results for each method and supports that our method is able to capture the garment shape, fit, and details the best.

For the sizing free-form question, participants again paid attention to different aspects of the image to make their decisions. Some focused on the length of the garment and garment fit (``[SizeGAN] has longer hem and has more looseness on the sides") to make their decisions, while others paid attention to model size. For this particular example, VITON-HD was often chosen as the largest size, but people included that it did not look real: ``[VITON-HD] looks to be the largest but its not the correct shirt."

\begin{figure}[tb]
  \centering \includegraphics[width=0.9\linewidth]{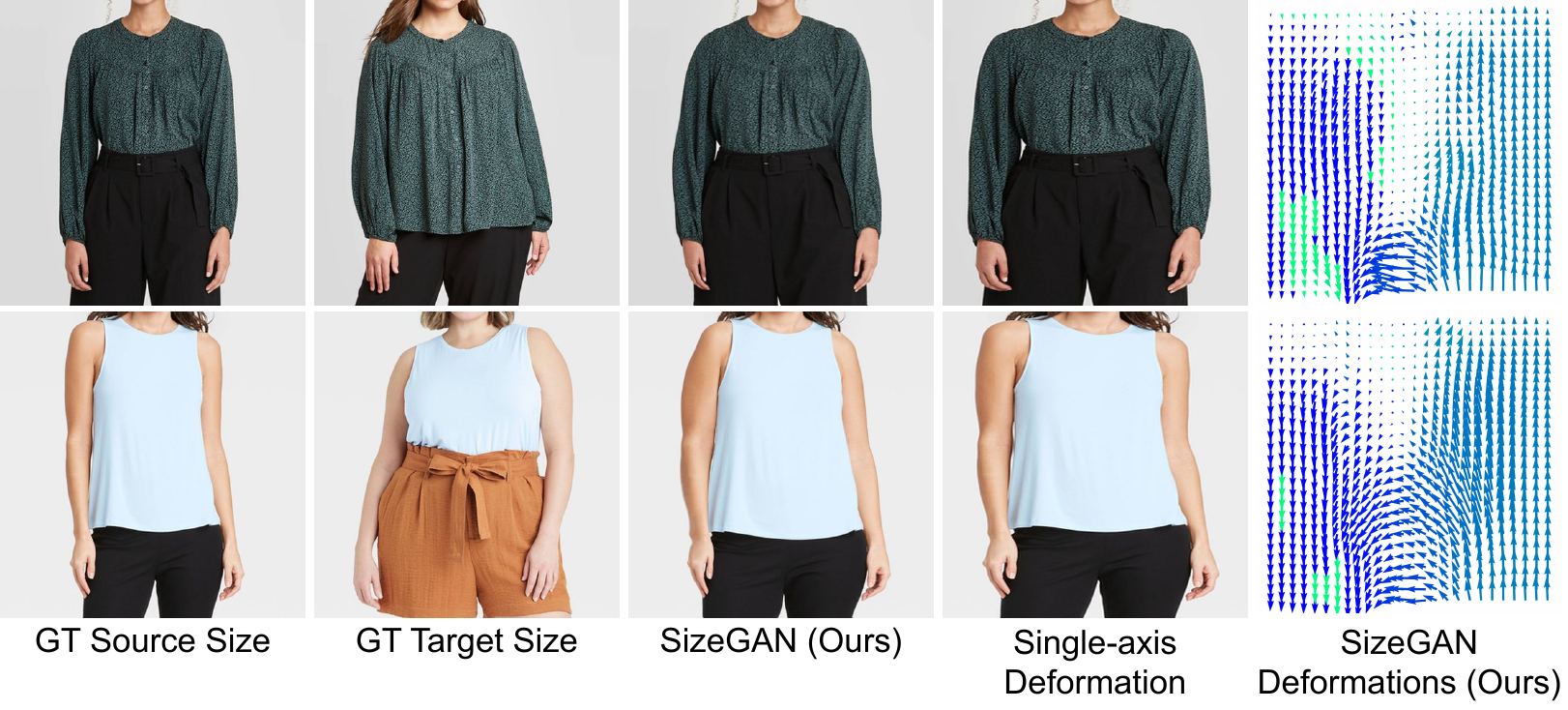} 
  \caption{Compared to the single-axis deformation baseline, SizeGAN learns complex deformation fields that result in body proportions more similar to ground truth. For better visualization, we use a stride of 20 to reduce the number of plotted points in each deformation field.
   }
   \label{fig:gt}
\end{figure}

\begin{figure*}[tb]
  \centering \includegraphics[width=0.6\linewidth]{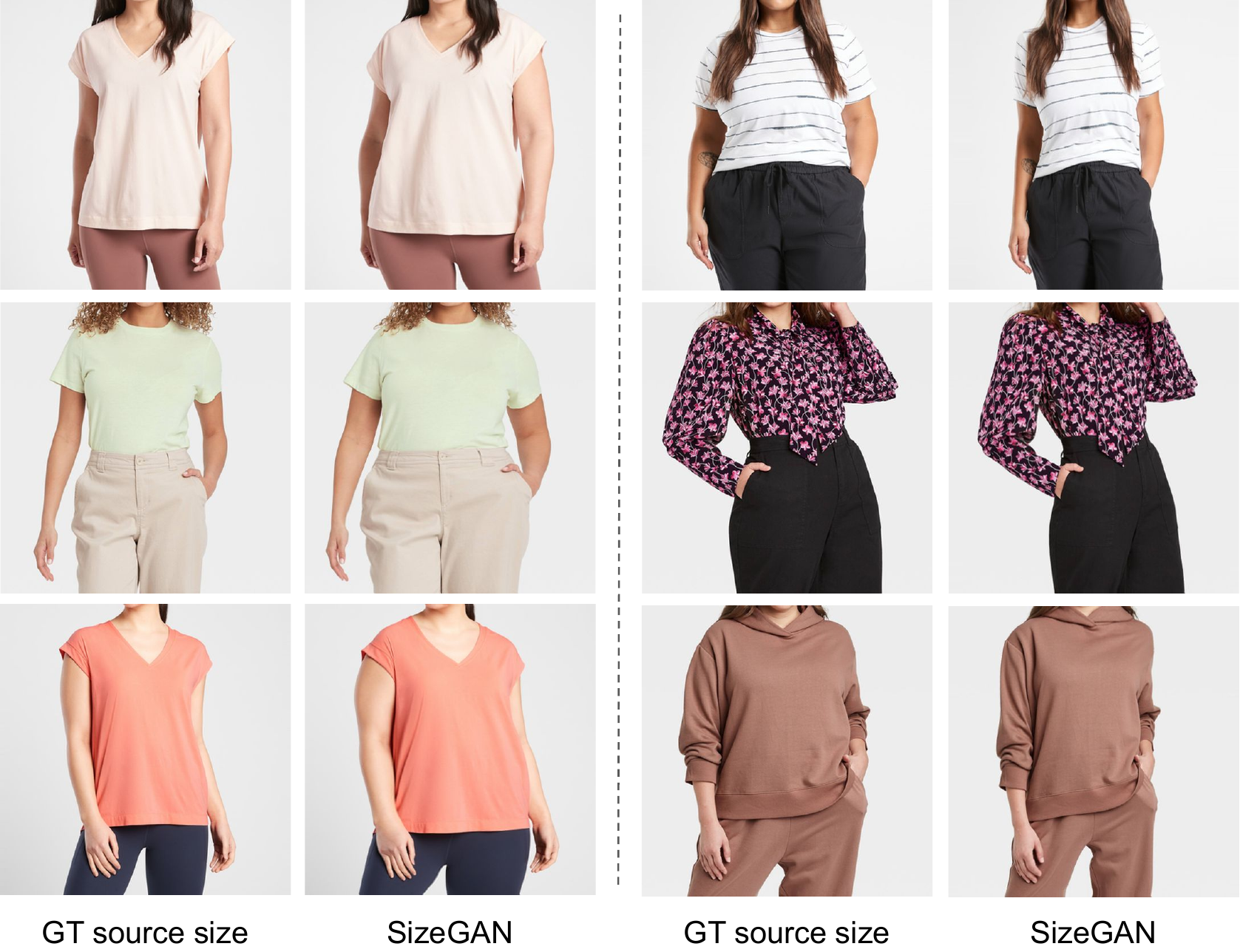}
   \caption{SizeGAN results on the test set from our new dataset. SizeGAN learns deformation fields that create natural, realistic images of garments in a new size. On the left, we show resizing from a smaller size to a plus size. On the right, we show resizing from a plus size to a small size.}
   \label{fig:ourDataset}
\end{figure*}

\subsection{Comparison with Single-axis Deformation Baseline} 
We ran an additional user study\footnote{Second user study link: https://forms.gle/YPxwA898YKazwywL6} on the in-distribution test set from our collected dataset. We compare 10 \textit{randomly} chosen SizeGAN test set results to the corresponding single-axis deformation baseline results. In this study, we evaluate size precision and realism by asking Amazon Mechanical Turk workers ``Which image looks more like a realistic plus sized model?''. Participants could choose image A, image B, or a tie. For three questions, we asked them to briefly explain their answer as a quality control. We also had one repeat question to ensure the participants' choices were consistent throughout the survey. We only counted the result from the first question (not the repeat question). We showed six real 1X plus size models in the introduction to calibrate the users. We had 280 usable responses from 28 participants. Of the responses that expressed a preference, 79\% preferred SizeGAN.

Figure \ref{fig:gt} shows test set examples of the single-axis deformation baseline and SizeGAN results in comparison with the ground truth models. It also shows deformation fields. SizeGAN learns to change the image in a realistic way, whereas the single axis deformation baseline creates unrealistic proportions. For example, the shoulders are squared off rather than being curved as in the ground truth images.  

\subsection{Qualitative Evaluation}
We show qualitative results for several images in the test split of our dataset in Figure~\ref{fig:ourDataset}. All of these result images were correctly classified by our pre-trained classifier as the target size. SizeGAN learns to change the model and garment size to not only match the target distribution, but also to appear natural and preserve the garment appearance. 

Figure~\ref{fig:ourDataset} (left) shows three examples of generating plus size models from smaller sizes. In addition to changing the size of the top, SizeGAN makes other changes that contribute to the realism of the image. For example, it changes the bare arms and the shape of the hips. It also maintains the 3D aspects of the image, e.g. folds of the shirt. 

Figure~\ref{fig:ourDataset} (right) shows three examples of generating small size models from larger sizes. These were generated using a SizeGAN model re-trained to deform larger sized tops to smaller sized tops. This demonstrates that the method works in both directions when trained on the appropriate data. The first two rows demonstrate that SizeGAN preserves patterns when resizing the garments. There is considerable variance in the source size models, and that variance is preserved by SizeGAN. For example, both the source and the SizeGAN models in row two of Figure~\ref{fig:ourDataset} are smaller than those in the first row of that figure.

\section{Limitations}
\begin{figure}[t]
 \begin{center}
    \includegraphics[width=0.8\linewidth]{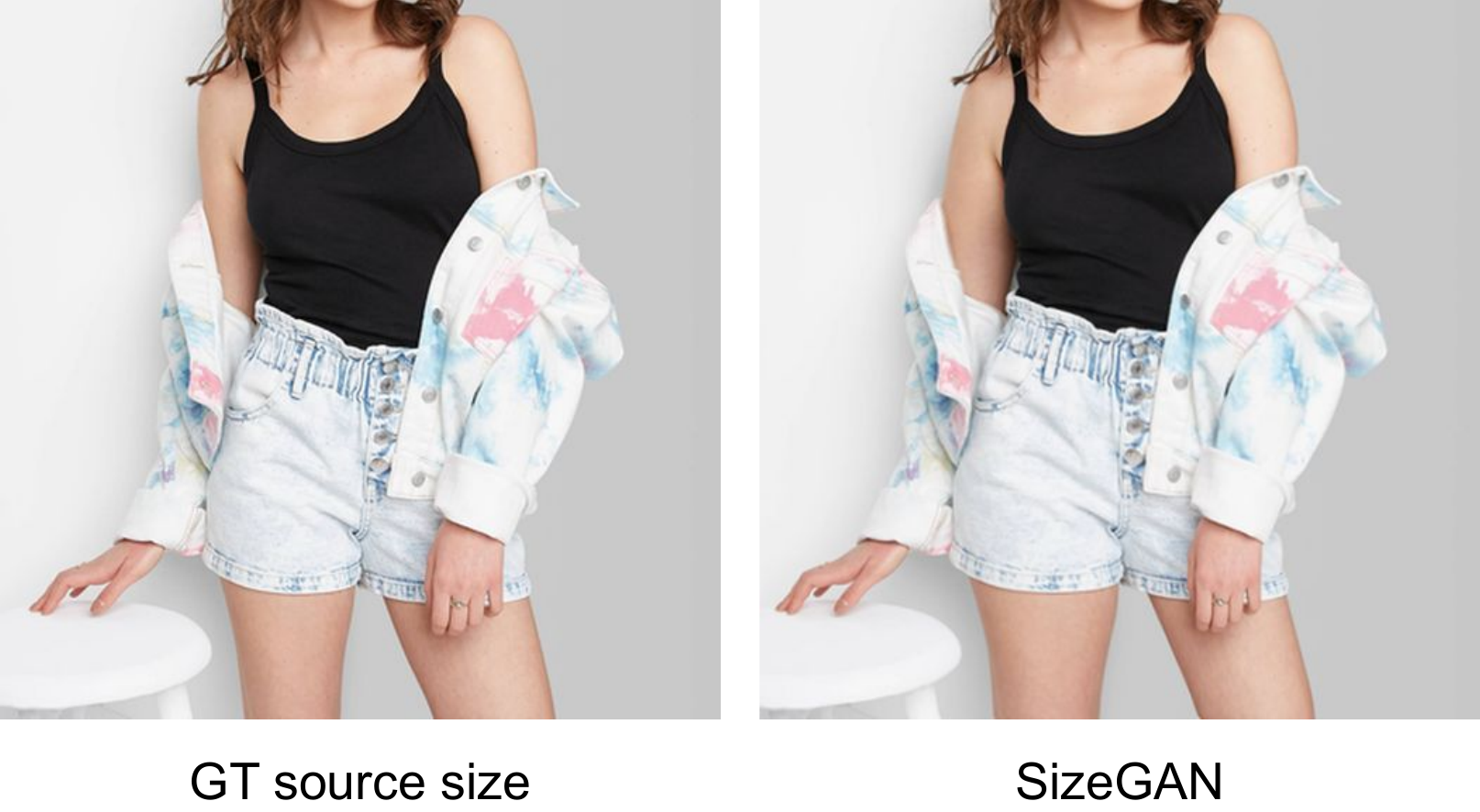}
    \end{center} 
    \caption{In cases where there is a busy background, several accessories, or a difficult pose, SizeGAN does not transform the model and garment correctly. In these cases, it often defaults to producing a small deformation field, leading to minor changes in the final image. }
  \label{fig:limitations}
\end{figure}

As shown by the baseline results, the resizing problem is non-trivial. Our method and dataset lay the ground work for addressing this important problem, but more work is needed.Our method does not generalize well to some images where there is a busy background, a difficult pose, or several accessories present (Figure~\ref{fig:limitations}). This is probably because of the limited amount of training data. Additionally, because of our limited training data, SizeGAN only has two target sizes. While SizeGAN realistically reshapes models and garments, we do not attempt to synthesize pattern repetition (e.g., adding additional stripe) or to synthesize new folds. Such enhancements could be complimentary to our deformation-based method. Our method aims to produce catalog images, and, unlike try-on methods, cannot be used to demonstrate what a garment would look like on specific individuals. However, SizeGAN could be used for training-time augmentation to improve try-on size generalization.

\section{Conclusion}
In this work, we propose SizeGAN, the first machine learning approach that addresses the \textit{size under-representation} problem. Online clothing stores often lack model size diversity, which can be frustrating for customers and can also lead to negative health consequences~\cite{body_exposure,model_body_size}. SizeGAN is able to change the size of a garment and model in an image to produce more diverse visualizations of a garment. We evaluate SizeGAN along three dimensions: realism, garment faithfulness, and size. Our first user study shows that SizeGAN outperforms current methods, and was vastly preferred for realism and garment faithfulness. We further show in a second user study that SizeGAN outperforms an alternative resizing approach, the single-axis deformation baseline. Lastly, we release the first paired clothing dataset with size labels, which we hope will inspire more work in this area.

{\small
\bibliographystyle{ieee_fullname}
\bibliography{main}
}

\section{Supplement}

\subsection{VITON Dataset}
We downloaded a random subset of 300 images from the VITON test dataset and removed images that were taken at an angle or had models in out of distribution poses (images cropped above mid-thigh or models facing to the side or backwards). To be fair to the baselines, we also removed images with matching tops and bottoms. SizeGAN keeps the bottoms in the original image, while the try-on methods only transfers the original shirt. For a matching track-suit, for example, SizeGAN results would keep the matching outfit, while the try-on results would have non-matching tops and bottoms. This could possibly effect the apparent realism of each result and unfairly bias the user study in favor of SizeGAN. After filtering, we had 168 testing images. We randomly sampled 15 images from this set to include in the user study.

\subsection{User Study Responses}
In the first study, participants were asked to rank the methods along three dimensions: realism, garment faithfulness, and size. The users saw each image labeled as ``A", ``B", or ``C". In the spreadsheet, we provide a mapping of which method each letter refers to in each question. 

Additionally, all user study participants provided free-form text justifications of their responses for the same three questions. Some users were asked to provide rank justifications for more than three questions. In the paper we focus on the free-form text responses from the three questions that all participants answered. Lastly, there were a few participants who had typos in one of their question responses (e.g. ``c,b,c" as their ranking). If the rest of their responses answered the questions and indicated they understood the task, we kept these survey responses. We did not include the answer with a typo when calculating the results though. 

In the second user study, the users saw each image labeled as ``A'' or ``B''. We provide a mapping of these labels to methods and all participant responses.

\begin{figure}[t]
  \begin{center}
  \includegraphics[width=\linewidth]{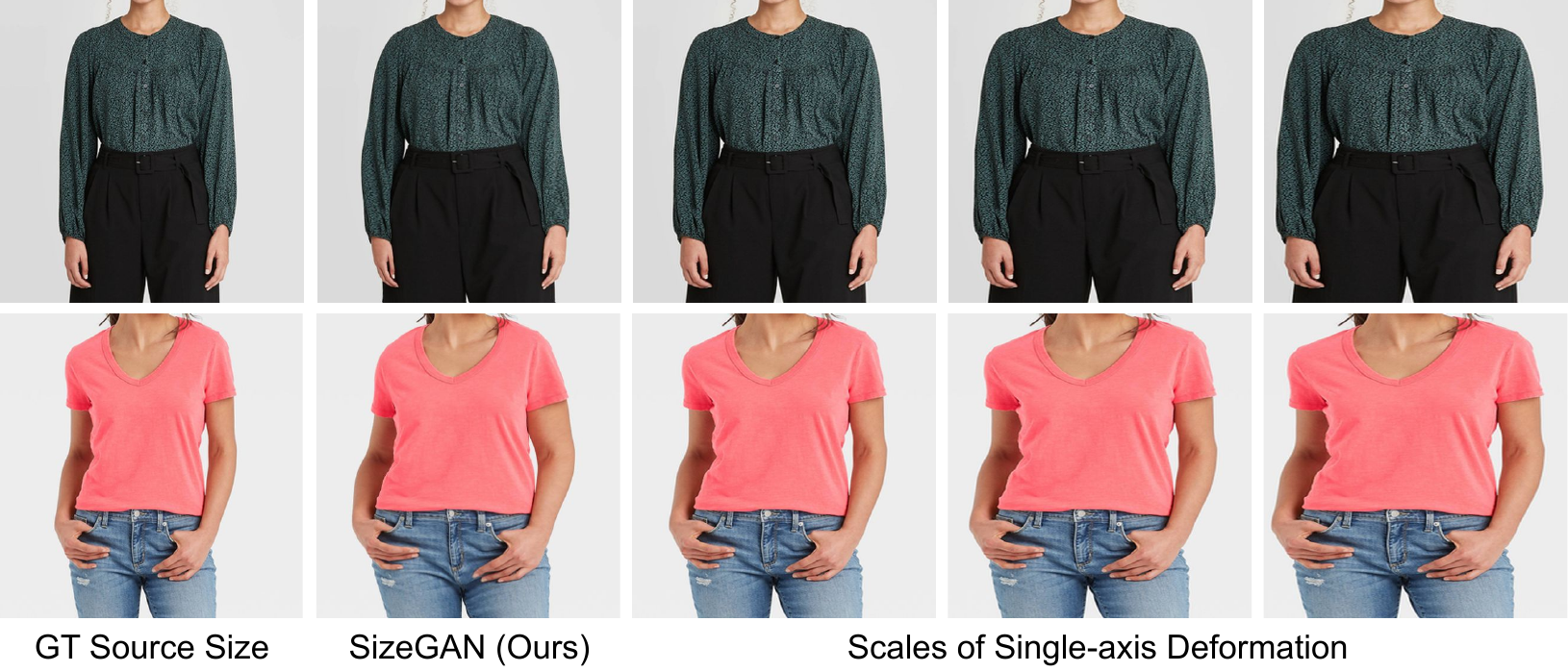}
  \end{center}  
   \caption{For the single-axis deformation approach, we take the average distance between hip pose keypoints in each size distribution in the training set and scale $I^A$ by this ratio (1.36) in the x direction to get $\tilde{I}^{B}$ (fourth column). We also show qualitative results for the single-axis deformation baseline for ratios 1.2 (third column) and 1.5 (fifth column). SizeGAN is able to match the target size naturally, whereas the single-axis deformation baseline does not have a good trade-off between correct size and realism.}
   \label{fig:scales}
\end{figure}

\subsection{Single-axis Deformation Baseline}
In the main paper, we show user study results comparing to the single-axis deformation baseline that uses a ratio of 1.36. We additionally show qualitative results in Figure \ref{fig:scales} for ratios, 1.2 and 1.5. The single-axis deformation baseline does not have a good trade-off between realism and the correct size. For the ratio 1.2, the garment size is either not close to the target size or has unrealistic proportions (e.g. shoulders are squared off). For the ratio 1.5, the image looks too stretched. SizeGAN is able to change the size of the model and garment in a realistic, natural way compared to the single-axis deformation baseline.

\end{document}